\documentclass[sigconf,natbib=true,nonacm]{acmart}

\AtBeginDocument{%
  }

\begin{document}

\title{Investigating Masking-based Data Generation in Language Models}

\author{Ed S. Ma}
\authornote{Independent project. Work in Progress. Contact: eddsma@outlook.com}
\email{eddsma@outlook.com}

\renewcommand{\shortauthors}{Ed S. Ma}

\begin{abstract}

The current era of natural language processing (NLP) has been defined by the prominence of pre-trained language models since the advent of BERT. A feature of BERT and models with similar architecture is the objective of masked language modeling, in which part of the input is intentionally masked and the model is trained to predict this piece of masked information. Data augmentation is a data-driven technique widely used in machine learning, including research areas like computer vision and natural language processing, to improve model performance by artificially augmenting the training data set by designated techniques. Masked language models (MLM), an essential training feature of BERT, have introduced a novel approach to perform effective pre-training on Transformer based models in natural language processing tasks. Recent studies have utilized masked language model to generate artificially augmented data for NLP downstream tasks. The experimental results show that Mask based data augmentation method provides a simple but efficient approach to improve the model performance. In this paper, we explore and discuss the broader utilization of these data augmentation methods based on MLM.

\end{abstract}


\maketitle

\section{Introduction}\label{sec-intro}

Pre-trained language models (PLMs) have revolutionized the field of natural language processing, with BERT architectures standing out for their innovative design and impressive performance. These Transformer \cite{vaswani-etal-2017-attention} based models, based largely on transformer architecture, use bi-directional representations to understand context, thus pushing the limits of previous uni-directional models. A distinctive feature of BERT and similar models is the goal of masked language modeling, in which part of the input is intentionally masked and the model is trained to predict this masked token. This strategy simulates a fuller understanding of the context and relationships between words, leading to a better understanding of language nuances and meaning. As a result, BERT-like models have found extensive application in a variety of NLP tasks, such as text classification, sentiment analysis, and question answering, significantly pushing the boundaries of what machines can understand and accomplish in the realm of human language.

NLP tasks requires a significant amount of high-quality annotated data for several critical reasons, primarily related to the inherent complexity of human language and the need for machine learning models to model it effectively understand, interpret and generate. Languages are extremely complicated and complex, with countless nuances, exceptions and rules. Consisting of morphological, syntactic, semantic, and pragmatic aspects, they require understanding not only of the words and phrases, but also of context, intent, and even cultural or social cues. In order for an NLP model to understand all of these elements and generate human-like text, it must learn from a variety of examples that show these characteristics in multiple different contexts. This is where high-quality annotated data comes into play. They provide the models with explicit labels or additional information that make it easier to understand the various features of the language.

The performance of machine learning models is highly dependent on the variety and amount of training data. Because these models learn by identifying patterns in the input data, a larger and more diverse data set allows the models to be exposed to a wider range of patterns and situations. This leads to better generalization ability and the models can process new inputs more effectively. For example, if you train an NLP model on annotated data from different domains like literature, science, law, social media, etc., it can understand and generate texts related to each of these domains. The quality of the annotated data is important. Poorly annotated data can mislead the model during training, resulting in suboptimal performance or even completely wrong outputs. Accurate annotations are fundamental to supervised learning as they serve as the basis for the model. They help models distinguish between different elements of language, understand the relationships between words, and understand the meaning and intent behind phrases or sentences. Therefore, high-quality annotated data plays a crucial role in training robust and reliable NLP models. It provides the rich, diverse, and concise input the models need to learn the complexities of the language, ensures their applicability in different domains and scenarios, and acts as an effective guide during the training process to optimize their performance.

It is often difficult and expensive to obtain quality annotated text data in large volume. Traditional and expensive way is to hire crowd workers with target language capability to annotate data. An example is Amazon Mechanical Turk (AMT). AMT is a crowdsourcing service that enables individuals and businesses to outsource tasks to a distributed workforce who can perform these tasks virtually. Based on requirements, workers (known as `Turkers`) will go through target data, manually annotate it as per instructions. Once a worker completes a Human Intelligence Task (HIT), you can review their work, approve or reject it based on the quality of the annotation, and then pay the worker. With its vast, diverse workforce, AMT is often used to create large, annotated datasets for NLP downstream tasks. However, this annotating method is common but very expensive. Therefore, researchers have started exploring annotating methods at cheaper costs. Some examples are methods that are based on distant supervision. Unlabeled dialogue corpora in the target domain can be easily curated from previous conversation transcripts or collected via crowdsourcing \cite{budzianowski-etal-2018-multiwoz,byrne-etal-2019-taskmaster} with no additional cost on human labour.

With the rise of BERT (Bidirectional Encoder Representations from Transformers) \cite{devlin-etal-2019-bert} models in NLP, there has been a significant leap forward in the field's capabilities and applications. BERT's bi-directional approach, where it considers the context from both left and right of a word, sets it apart from previous models. The experiments on several NLP benchmark shows that it allows a more accurate understanding of the meaning of a word within its context, leading to improved performance in a variety of tasks such as sentiment analysis, named entity recognition, question answering, and others. There are several data augmentation methods that utilize these pre-trained language models in unsupervised manners, with no human annotation needed.

\section{Related Work}\label{sec-related}
In this section, we will introduce pre-trained language models and recent data augmentation methods including data augmentation methods with MLM.

\subsection{Pre-trained Language Models} \label{subsec-pretrained}

Pre-trained language models such as BERT \cite{devlin-etal-2019-bert}, RoBERTa \cite{liu-2019-roberta}, XLNet \cite{yang-etal-2019-xlnet}, BART \cite{lewis-etal-2020-bart}, and T5 \cite{raffel-etal-2020-t5} have revolutionized the field of Natural Language Processing. These models rely on transformer-based architectures and unsupervised learning, and they are trained on large text corpora to learn useful representations of language. They can then be fine-tuned on specific tasks with smaller amounts of task-specific data. The objective is masked language modeling is essential and prominent in these models,

\begin{figure}[tb]
	\centering
	\includegraphics[width=\linewidth]{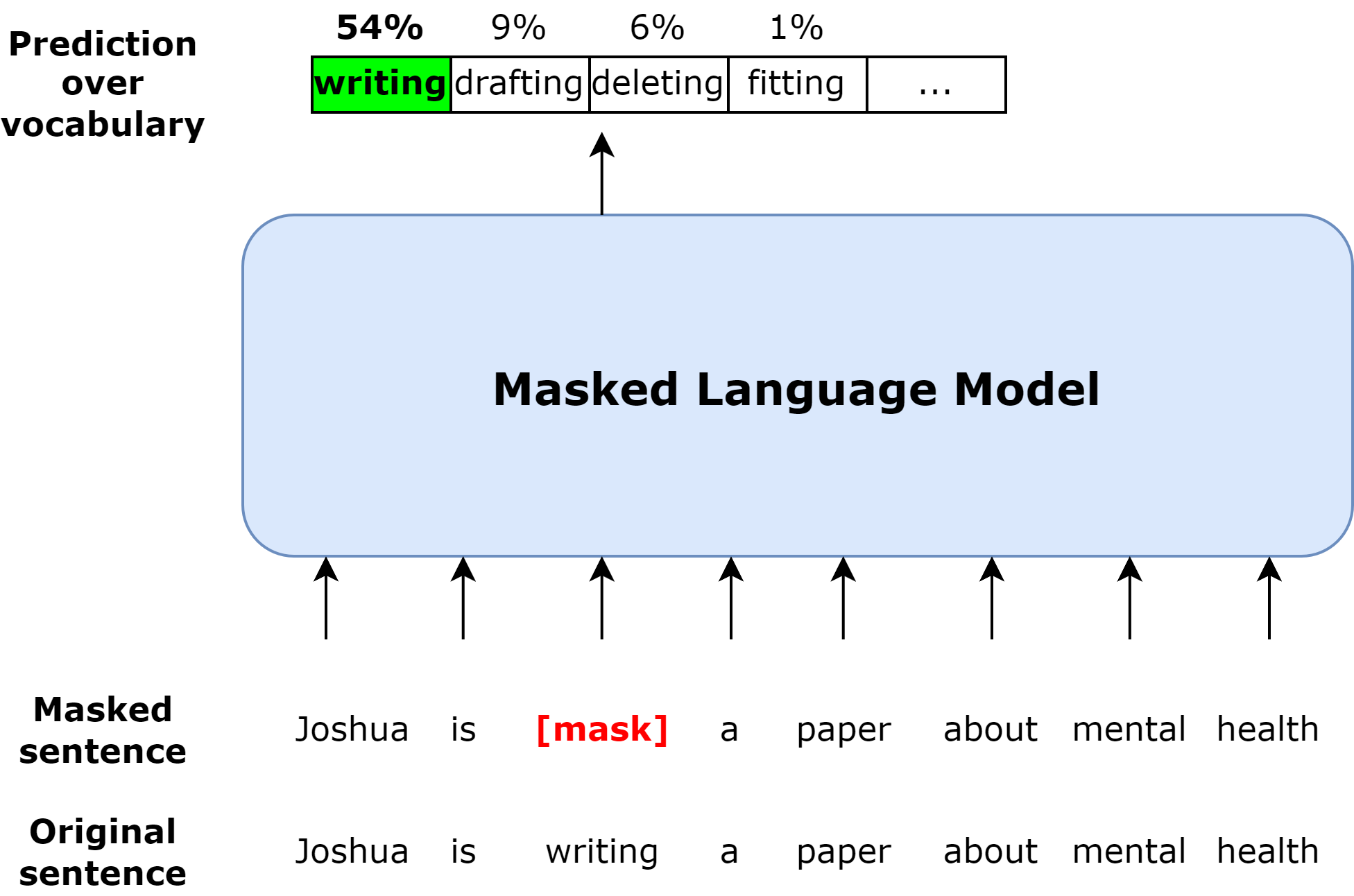}
	\caption{Visual illustration of MLM in BERT. During pre-training, some tokens (about 10 -- 15 percent) are masked randomly (only one token is masked in this example) and the model is trained to predict the correct words for all masked tokens. The model will be updated based on the probability distribution over vocabulary through MLM loss functions and backprobagation. Numbers are for illustration purposes instead of real model output.}
	\label{fig/Fig-bert-mlm}
\end{figure}

BERT \cite{devlin-etal-2019-bert}, Bidirectional Encoder Representations from Transformers, was a pioneering pre-trained language model in the field. It is trained using a masked language modeling objective where some percentage of the input data is masked or hidden during training, and the model is trained to predict these masked words based on the context provided by the unmasked words. BERT's MLM approach enables the model to learn a deep, bidirectional representation of the sentence, rather than just predicting future words in the text like traditional language models. This improves the model's understanding of the context and the semantic relationships between words, leading to significant improvements on a wide range of NLP tasks. We show a visual illustration of MLM in BERT in Figure \ref{fig/Fig-bert-mlm}. Since BERT is a bi-directional model, the probability of predicting the correct masked word is dependent on known surrounding words in the context. RoBERTa \cite{liu-2019-roberta}, a variant of BERT developed by Meta, modifies the training process and hyperparameters of BERT, but still utilizes the MLM objective. RoBERTa extends the training time, uses larger mini-batches and learning rates, and removes the next sentence prediction objective that BERT had. These tweaks led to improved performance, demonstrating the impact of a well-thought-out training strategy.

XLNet \cite{yang-etal-2019-xlnet} is a another transformer-based model, combining the advantages of BERT with autoregressive language models. It introduces a permutation-based training objective that allows it to learn the dependency of words on both their preceding and succeeding context, mitigating some of the limitations of BERT's MLM approach where the predicted words don't have an impact on each other.

T5 \cite{raffel-etal-2020-t5} Text-to-Text Transfer Transformer takes a unique approach to its pre-training objective. Instead of employing the usual masked language modeling, T5 treats every NLP problem as a text generation task. This is achieved by reframing tasks into a unified text-to-text format where every task, be it translation, summarization, or question answering, becomes a matter of generating the target text from the source text. During pre-training, T5 uses a denoising autoencoder-style objective where random spans of text are masked out and the model learns to reconstruct the original text. T5's approach simplifies the application of the model to a wide variety of tasks, as it doesn't require task-specific model architectures. Instead, differences between tasks are captured purely in the text input and output formats. As for performance, T5 has shown state-of-the-art results on a number of benchmark datasets across different NLP tasks. Its scalability, coupled with the general text-to-text approach, makes T5 a highly flexible and powerful tool for NLP tasks, showcasing the potential of this unified framework.

BART \cite{lewis-etal-2020-bart} Bidirectional and Auto-Regressive Transformers brings a distinctive pre-training objective to the table. BART combines the best of both worlds: the bidirectional context learning from BERT and the sequence-to-sequence nature of models like GPT.
During pre-training, BART corrupts the text by applying a noising function (i.e., by masking out random spans of text, similar to T5 \cite{raffel-etal-2020-t5}), and then learns how to reconstruct the original uncorrupted text.
This objective is different from traditional masked language modeling as it forces the model to build an understanding of the entire sentence structure and not just predict masked tokens independently.
BART has proven to be highly effective. It has achieved state-of-the-art results on a variety of benchmark datasets, including the CNN/Daily \cite{nallapati-etal-2016-abstractive} Mail summarization task, the SQuAD question answering benchmark \cite{rajpurkar-etal-2016-squad,rajpurkar-etal-2018-know}, and others.
By combining the advantages of bidirectional and auto-regressive transformers, BART has shown a robust ability to handle a wide range of NLP tasks, highlighting the power of sequence-to-sequence and denoising autoencoder-style pre-training.

Pre-trained language models have had a profound impact on NLP, and the Masked Language Modeling objective has been a critical part of their success. By learning to predict missing words in a sentence, models are able to gain a deeper understanding of language context and semantics. This has resulted in significant performance improvements across a wide range of NLP tasks, and it continues to drive progress in the field.

Masked language modeling plays a crucial role in these notable pre-trained language models. By learning to fill in the masked words, language models gain a deep understanding of syntax, semantics, and contextual relationships. Masked language modeling serves as a self-supervised learning task that allows models to learn from vast amounts of unlabeled text data. This technique allows pre-trained language models to capture the nuances of language and generate coherent and contextual responses when presented with incomplete or ambiguous input. By incorporating masked language modeling, pre-trained language models have revolutionized various natural language processing tasks such as text generation, sentiment analysis, machine translation, and question answering.

\subsection{Data Augmentation} \label{subsec-data-aug}

Data augmentation (DA) is a technique used to generate additional training data in situations where there is a lack of sufficient data. It involves various methods, ranging from simple rule-based techniques to more advanced learnable generation-based methods. The primary objective of data augmentation is to ensure that the generated data is valid for the given task and belongs to the same distribution as the original data \cite{raille2020fast}. This validity is determined by factors such as similar semantics in machine translation or the same labels in text classification as the original data. Additionally, augmented data should also exhibit diversity to enhance the generalization of models for subsequent tasks. The diversity of augmented data can be achieved through three categories of data augmentation methods: paraphrasing, noising, and sampling.

\noindent\textbf{Categorization}

In order to ensure validity, augmented data should also possess diversity to enhance the generalization of models in subsequent tasks. This diversity refers to the range of variations within the augmented data. \cite{li-2022-data} introduces categorization of data augmentation methods based on the diversity of resulting augmented data. Paraphrasing-based methods generate augmented data that maintains a limited semantic difference from the original data by making proper and controlled changes to sentences. The goal is to produce augmented data that conveys very similar information as the original. Noising-based methods introduce discrete or continuous noise to the data while ensuring its validity. The objective of these methods is to enhance the model's robustness. Sampling-based methods, on the other hand, understand the data distributions and generate novel data samples from within these distributions. These methods output more diverse data and cater to a wider range of requirements in downstream tasks, leveraging artificial heuristics and trained models.

Paraphrases are commonly observed in natural language, initiated from early works in NLP \cite{barzilay-mckeown-2001-extracting,madnani-dorr-2010-generating}, serve as alternative expressions to convey identical information as the original form. Consequently, utilizing paraphrasing as a technique for data augmentation is a fitting approach. Paraphrasing encompasses multiple levels, encompassing lexical paraphrasing, phrase paraphrasing, and sentence paraphrasing.

\noindent\textbf{Basic word-/sentence-level operations}

Various basic word-level and sentence-level operations can be employed to augment text data, including insertion, deletion, substitution, and swapping. Insertion, additional words or phrases are inserted into the original text. By inserting different linguistic elements, such as synonyms or context-relevant terms, the extended data can simulate different writing styles and increase the diversity of the data set. \cite{wei-zou-2019-eda,peng-2021b-data} This variation can allow the model to learn better and generalize to different text types. Deletion, on the other hand, is the removal of specific words or phrases from the original text. This technique helps the model focus on the most important information by eliminating non-essential elements. This operation has been employed in various works such as \cite{wei-zou-2019-eda,yu-etal-2019-hierarchical,rastogi2020achieve} By training on augmented data with strategically deleted content, the model can become more resilient to noise and less dependent on specific phrases or words. Substitution is another powerful data augmentation technique. Words are replaced by their synonyms or similar terms. By introducing variations in vocabulary, the model can learn to recognize and understand different word choices and expand its language skills. \cite{xie-etal-2017-data-noising,wang-etal-2018-switchout,daval-frerot-weis-2020-wmd,lowell-etal-2021-unsupervised,regina2021text} This technique is particularly useful for improving the model's ability to deal with out-of-vocabulary words and to adapt to different linguistic expressions. Swapping is an augmentation technique that rearranges the order of words or phrases within a sentence. By rearranging the sequence of linguistic elements, the model can learn to understand different sentence structures and syntactic variations. \cite{wei-zou-2019-eda,zhang-etal-2020-on-data,longpre-etal-2020-effective,rastogi2020achieve,howard-etal-2022-neurocounterfactuals} This increases the model's flexibility and understanding of different sentence patterns, making it more effective in tackling sentence reordering or paraphrasing tasks.

\noindent\textbf{PLM backed methods}

As mentioned in Section \ref{subsec-pretrained}, the success of deep pre-trained language models in recent years can be attributed to their ability to acquire extensive linguistic knowledge through pre-training. As a result, these sophisticated pre-trained models have naturally become valuable resources for data augmentation purposes.

Auto-regressive models have been explored in generating augmented data. For instance, \cite{peng-2021b-data} utilizes the pre-trained GPT based models to generate utterances and dialogue acts, respectively, ensuring data quality through filtering. \cite{Abonizio-2020-pretrained} applied DistilBERT on original sentences to generate synthetic sentences. Transformer decoder based model GPT-2 \cite{radford2019language}, in particular, has gained popularity as a model for generating augmented data. \cite{zhang-etal-2020-on-data} employed GPT-2 to generate significantly diversified augmented data for extreme multi-label classification.
Another GPT-2 backed data augmentation method LAMBDA is introduced in \cite{anaby-tavor-2020-do-not}. They utilize GPT-2, which is pre-trained and fine-tuned on the training set, to generate labeled augmented sentences. These augmented sentences are subsequently filtered using a classifier to maintain data quality. \cite{kumar-etal-2020-data} applies a similar method without the use of a filtering classifier.
\cite{quteineh-etal-2020-textual} employed a label-conditioned GPT-2 model to generate augmented data. \cite{tarjan2020deep} utilized GPT-2 to generate augmented data and subsequently performed tokenization and splited them into subwords derived statistically, thereby avoiding vocabulary explosion in morphologically rich languages.

To obtain augmented data, some studies employ masked language models. For instance, \cite{ng-etal-2020-ssmba} utilizes a masked language model to create both a corruption model and a reconstruction model. The model comprises a BERT-base pre-trained language model and a classification layer. The BERT parameters are additionally pre-trained on a vast Reddit dataset, ensuring a large-scale training process. The process involves generating data points that are initially distant from the original data manifold using the corruption model, and subsequently using the reconstruction model to bring those data points back to the original data manifold, resulting in the final augmented data. Such decoding sampling methods have also been utilized in \cite{gao-etal-2022-mask} which first masks tokens or spans then fills in replaced tokens by pre-trained language models.

Obtaining unlabeled raw data can be relatively easy in certain scenarios, presenting an opportunity to convert such data into valid, usable data and significantly augment the overall dataset.
\cite{thakur-etal-2021-augmented} introduce a data augmentation approach involving fine-tuning BERT on the original data, followed by utilizing the fine-tuned BERT to label unlabeled sentence pairs. These augmented data, along with the gold data, are then combined to train SBERT \cite{reimers-gurevych-2019-sbert}.
Data distillation has been utilized as part of the self-training process \cite{miao-etal-2020-twitter}, where the label of unlabeled data is determined using an iteratively updated teacher model. On question answering tasks, \cite{yang-etal-2021-neural-retrieval} applies a similar self-training method using a cross-attention-based teacher model to determine the label for each QA pair.
SentAugment \cite{du-etal-2021-self}, a data augmentation method that leverages task-specific query embeddings from labeled data to retrieve sentences from a vast bank of billions of unlabeled sentences obtained from web crawling.
In some cases, existing models from other tasks are directly transferred to generate pseudo-parallel corpora. \cite{montella-etal-2020-denoising} leverages Wikipedia to access a large volume of sentences and utilizes external information extraction package to extract triplets from these Wikipedia sentences. Recognizing BERT's proficiency in object-property (OP) relationship prediction and object-affordance (OA) relationship prediction, \cite{zhao-etal-2020-learning} directly employs a fine-tuned BERT model to predict the labels of the two samples. \cite{wei-etal-2022-learning} explore data augmentation on semantically-preserved continuous space using Transformers.

\noindent\textbf{Unsupervised methods}

Unsupervised methods for text data augmentation leverage unlabeled data to enhance the quantity and quality of training data without relying on explicit or extra annotations or labels. Some unsupervised methods have been mentioned under \textit{word-/sentence-level operations}, here we discuss methods not falling into those categories. These methods aim to harness the inherent structure, pre-training objectives and patterns present in the unlabeled data to generate additional training examples.

''Mixup'' methods take a different approach to data augmentation by utilizing intermediate embeddings instead of generating augmented samples in natural language text. These methods involve sampling data in vector space (manifold) based on existing data, potentially resulting in samples with different labels than the original data. A prominent feature of this group of methods is no additional human annotation is required to construct new training samples, and resulting labels are automatically generated after the ''mixing'' process. The concept of Mixup was initially introduced in computer vision works \cite{zhang2018mixup}. Mixup has found wide application in numerous recent studies. Building upon this work, \cite{Guo2019AugmentingDW} proposed two variations of Mixup specifically for sentence classification, performing sample interpolation in the word embedding space and interpolating the hidden states of sentence encoders. Both variations generate interpolated samples as augmented data. \cite{wu-etal-2022-text} propose a data augmentation method integrating Mixup strategy with MLM that involves converting a sentence from its one-hot representation to a controllable smoothed representation. \cite{kong-etal-2022-dropmix} introduce a framework for saliency map-informed textual data augmentation and regularization, which combines Dropout and Mixup techniques to address the issue of overfitting in text learning.

Mixup methods for text classification are explored in \cite{sun-etal-2020-mixup,si-etal-2021-better}. \cite{sun-etal-2020-mixup} introduces Transformer based Mixup method, a combination of Mixup and transformer-based pre-trained architecture, which is evaluated on text classification datasets. \cite{chen2020local} incorporates Mixup into Named Entity Recognition (NER) to improve performance in NER tasks. \cite{chen-etal-2020-mixtext} introduce MixText, which creates augmented training samples by interpolating text in hidden space. The method performing mixing of hidden representations of examples in different Transformer layers. The resulting labels are determined by mixing samples.

Machine translation has emerged as a popular method for data augmentation, leveraging its ability to naturally paraphrase text. With the advancements in machine translation models, this technique has found widespread application across various tasks. Researchers have explored off-the-shelf machine translation models on data augmentation with no additional human effort on annotations.
Back-translation, involves translating the original text into other languages and then translating it back to the original language to generate augmented text. Unlike word-level methods, back-translation doe not merely replace individual words, but rather rewrites the entire sentence in a generated manner.
\cite{wei2018fast,xie-etal-2020-unsupervised,fabbri-etal-2021-improving} utilize English-French translation models, in both directions (known as round-trip translations), to perform back-translation on each sentence and generate paraphrases. Lowell also incorporates this method as one of the unsupervised data augmentation techniques. Additionally, Zhang employs back-translation to obtain formal expressions of the original data in the context of style transfer tasks.

Building upon the concept of vanilla back-translation, several studies have introduced additional features and techniques to enhance its effectiveness. In \cite{zhang-etal-2020-parallel}, a discriminator is employed to filter the sentences obtained through back-translation. By applying this filtering mechanism, the quality of the augmented data is significantly improved, as only sentences surpassing a certain threshold are retained. \cite{nugent-etal-2021-flexible} explore various softmax temperature settings to ensure diversity in the augmented data while preserving semantic meaning. By carefully adjusting the temperature, they achieve a balance between generating diverse examples and maintaining the original intent. Overall, these additional features and techniques bring improvements to the vanilla back-translation method, ensuring diversity and enhancing the quality of the augmented data.

\begin{figure}[tb]
	\centering
	\includegraphics[width=\linewidth]{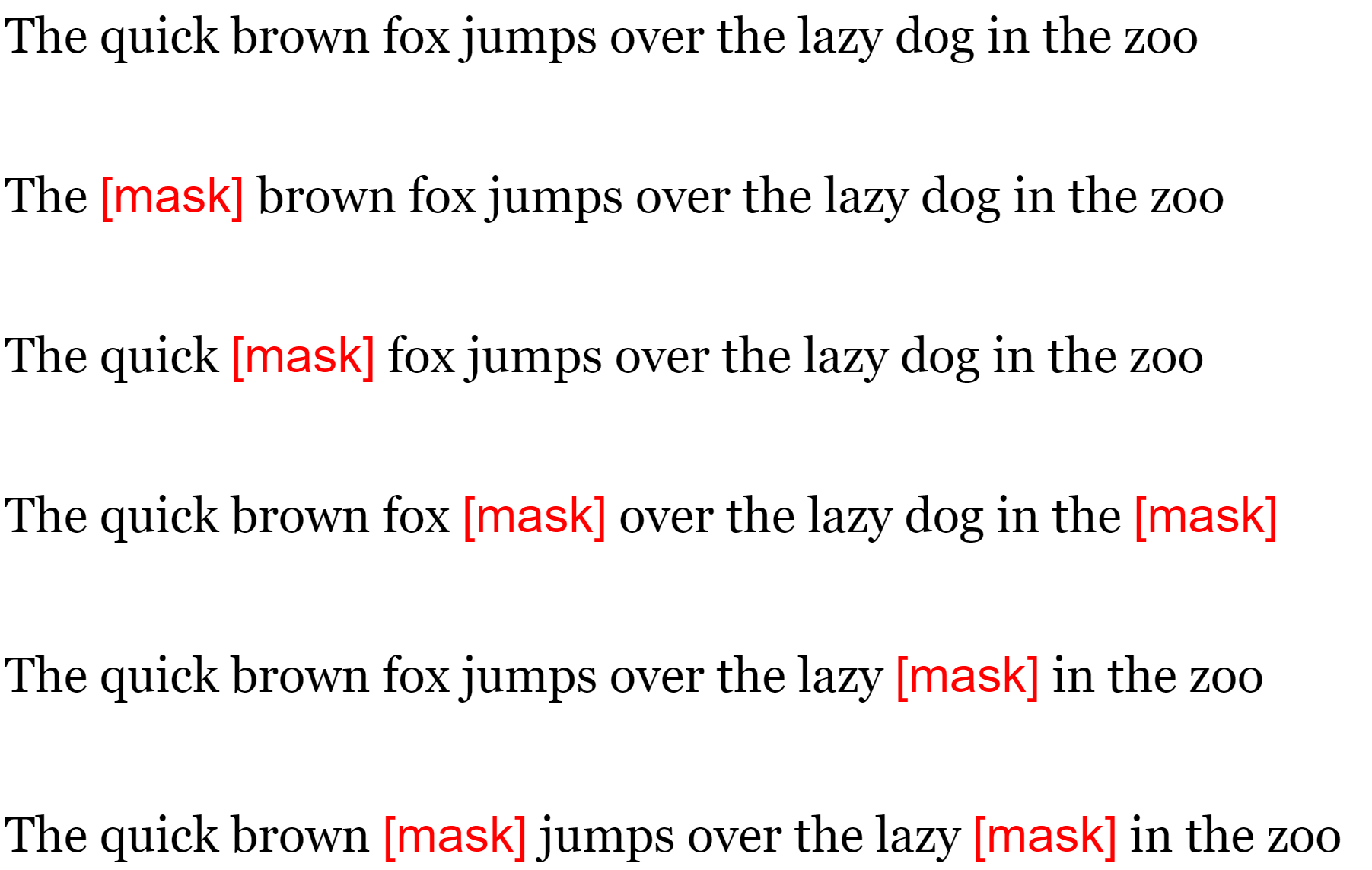}
	\caption{Example of augmented samples with mask tokens. The top first sentence is the original sentence without any augmentation (The quick brown fox jumps over the lazy dog in the zoo). The rest five sentences are augmented sentences obtained by data augmentation with mask tokens in which an original word in the original sentence is replaced with a MASK token.}
	\label{fig/Fig-mask-aug-example}
\end{figure}

\noindent\textbf{DA with Mask tokens}

Mask tokens are a critical component of pre-trained language models like BERT \cite{devlin-etal-2019-bert}. With BERT architecture, a small percentage of the input tokens are randomly masked during pre-training. These masked tokens are usually replaced with a special [MASK] token. We present a visual example of MLM in Figure \ref{fig/Fig-mask-aug-example} for ease of understanding. Using mask tokens in pre-trained language models such as BERT enables effective language representation learning and allows the models to capture the contextual relationships between words and phrases, resulting in improved performance on various NLP tasks. As a pre-training strategy, mask tokens are seen by the model during pre-training process. Mask tokens are pre-trained to preserve contextualized information (like other tokens) but not actually encountered in test-time examples. Intuitively, mask token can be used in sentences like other words without injecting undesired signals in sentences under the training scheme of BERT.

Word replacement and back-translation have been demonstrated to be effective unsupervised data augmentation methods for short written text classification in previous studies \cite{wei-zou-2019-eda,xie-etal-2020-unsupervised}. However, the effectiveness of these augmentation methods is diminished when applied to pre-trained models \cite{shleifer2019low}. Additionally, the applicability of backt-ranslation is limited in our scenario, given that translating multi-turn dialogue is considerably more challenging compared to short text.
Recent work \cite{yavuz-etal-2020-simple} has examined the effectiveness of replacing a portion of original word tokens with [MASK] tokens as a data augmentation strategy in NLP. This technique allows for the generation of extended examples in which certain words are escaped, providing the model with additional training instances. \cite{yavuz-etal-2020-simple} introduce \textsc{MaskAugment}, a controllable data augmentation that augments text input by leveraging the pre-trained mask token from BERT model on the task of dialog act tagging. \cite{lin-2022-semi} extend this idea and propose a semi-supervised bootstrapping training method for dialog breakdown detection tasks. Both works utilize teacher-student learning scheme on samples with different probabilities of mask token replacement. By leveraging the pre-trained language model's ability to understand and contextualize mask tokens, this extension method introduces diversity and variation into the training data without introducing unwanted bias or noise. This DA approach has been shown to increase model robustness, improve generalization to invisible data, and increase performance on various downstream NLP tasks. 

Compared to the aforementioned DA methods, mask data augmentation appears to be better suited for pre-trained language models with MLM training objectives and demonstrate effectiveness in dialog tasks. This method is intuitively simple and controllable as it only considers the position and probability of replacement. By maintaining a low probability of masking \cite{yavuz-etal-2020-simple,lin-2022-semi}, the sentences augmented with mask tokens can retain the original meaning while undergoing natural changes.

\noindent\textbf{DA with adversarial training}

NLP data augmentation with adversarial training has emerged as a promising technique to improve the robustness and performance of NLP models. Adversarial training involves generating adversarial examples or perturbations that aim to deceive the model while preserving the original meaning or intent of the text. In the context of NLP, these adversarial examples can be created by applying various techniques but not limited to synonym substitution, word deletion, or sentence modification to the original text.

\cite{cheng-etal-2020-advaug} constructs adversarial samples based on the original samples following \cite{cheng-etal-2019-robust} and then applies two Mixup strategies. \cite{qu2021coda} combines back-translation with adversarial training. \cite{morris-etal-2020-textattack} introduce an off-the-shelf framework for developing data augmentation with adversarial attacks covering various attack recipes from literature\cite{li-etal-2019-textbugger,jin-2020-bert-robust}. This integration allows them to synthesize augmented examples that are both diverse and informative, incorporating multiple transformations to enrich the data. The goal of adversarial data augmentation is to expose the model to challenging and diverse examples that mimic real-world linguistic variations and potential attacks. By incorporating these adversarial examples into the training data, NLP models can learn to handle and generalize better to such variations, making them more robust and reliable in real-world scenarios. Adversarial training not only helps models overcome the limitations of overfitting but also enhances their ability to handle noisy or adversarial inputs.

\section{Discussion}
\subsection{Non-MLM Pre-trained Language Models}
Nowadays, we are witnessing the rapid emergence of a diverse array of pre-trained language models that go beyond the traditional training objective of masked language modeling. Prominent examples include models such as T5 \cite{raffel-etal-2020-t5}, Flan \cite{chung2022scaling} and GPT \cite{radford2018improving,radford2019language}. T5/Flan-T5 deviate from the conventional approach of masked language modeling where individual tokens are masked. Instead, these models employ a different masking and prediction strategy that operates at the level of text spans. In this strategy, specific spans of text are masked, and the models are trained to predict the correct content within those masked spans. These models are designed with multifaceted training objectives aimed at enhancing different aspects of language understanding and generation. In addition to incorporating masked language modeling, these models can also integrate various other objectives such as sequence classification, document retrieval, question-answering, and machine translation. By encompassing this diverse range of training objectives, pre-trained language models gain the ability to capture a comprehensive understanding of linguistic properties, semantic relationships, and contextual nuances. This advancement in training objectives not only enhances the versatility and context-awareness of these language models but also facilitates their application across a wide spectrum of natural language processing tasks and domains. As ongoing research and innovation in this field continue to unfold, we can expect the emergence of even more advanced pre-trained language models, leading to significant advancements in natural language understanding and generation across diverse domains. 
Incorporating mask token-based data augmentation into the training pipeline of models like T5 and GPT is potentially feasible and explorable.

\subsection{Emerging Large Language Models}
The presence of large language models (LLM) like GPT-3 family models \cite{brown-etal-2020-language,ouyang2022training,openai-2022-chat} and Llama \cite{touvron2023llama} have brought both challenges and significant impacts on data augmentation methods.

One challenge is the scalability and computational demands that come with data proliferation. Large language models require a large amount of training data to achieve their impressive performance. However, generating extended data at the same scale can be computationally intensive and slow, requiring additional model processing and inference. On the other hand, the influence of large language models on data augmentation is significant. These models have extensive linguistic knowledge and can generate high-quality synthetic examples using techniques such as conditional generation or controlled generation. Consequently, data augmentation techniques can benefit from integrating these models to generate diverse and contextually relevant augmented data.

Additionally, large language models can serve as powerful tools for data enrichment. By optimizing or adapting pre-trained models like GPT-3 to specific downstream tasks, they can be used to generate advanced data tailored to the target task. This approach allows models to learn from extended data and potentially improve their performance on the given task.

Moreover, large language models can also be used for data enrichment by leveraging their embeds or contextual features. These embeddings can provide valuable information about relationships and similarities between text samples and facilitate the development of new data enhancement strategies that preserve semantic and syntactic properties.

\bibliographystyle{acm-ref-format}
\bibliography{reference}

% \appendix

% \section{Research Methods}

\end{document}